\documentclass{article}
\usepackage{IEEEtrantools}

\usepackage{spconf}
\usepackage{amssymb}
\usepackage{amsmath}
\usepackage{mathtools}
\usepackage{bm}
\usepackage{microtype}
\usepackage{siunitx}

\usepackage{xcolor}
\usepackage{graphicx}

\usepackage{caption}
\usepackage{subcaption}
\usepackage[export]{adjustbox}

\usepackage{cite}
\usepackage{url}

\usepackage{array}
\usepackage{booktabs}
\usepackage{mdframed}

\title{LARGE DEVIATIONS FOR CLASSIFICATION PERFORMANCE ANALYSIS \\ OF MACHINE LEARNING SYSTEMS}
\name{P.~Braca,$^*$ L.~M.~Millefiori,$^*$ A.~Aubry,$^\dagger$ S.~Marano,$^\ddagger$ A.~De~Maio,$^\dagger$ P.~Willett$^\mathsection$}
\address{$^*$ NATO STO Centre for Maritime Research and Experimentation (CMRE), La Spezia, Italy \\
$^\dagger$ University of Naples Federico II, Italy\\
$^\ddagger$ University of Salerno, Italy\\
$^\mathsection$ University of Connecticut, USA}
\begin{document}
\maketitle
\begin{abstract}
We study the performance of machine learning binary classification techniques in terms of error probabilities. 
The statistical test is based on the Data-Driven Decision Function (D3F), learned in the training phase, i.e., what is thresholded before the final binary decision is made.
Based on large deviations theory, we show that under appropriate conditions the classification error probabilities vanish exponentially, as $\sim \exp\left(-n\,I + o(n) \right)$, where $I$ is the error rate and $n$ is the number of observations available for testing.
We also propose two different approximations for the error probability curves, one based on a refined asymptotic formula (often referred to as exact asymptotics), and another one based on the central limit theorem. The theoretical findings are finally tested using the popular MNIST dataset.  
\end{abstract}
\section{INTRODUCTION}
\label{sec:intro}

A fundamental problem addressed by Machine Learning (ML) techniques, which spans several research and application fields, is to discover intricate structures in large datasets~\cite{lecun2015deep}.  %
Modern ML techniques based on Artificial Neural Networks (ANNs) are able to learn very complex functions, and in many contexts ANN methods represent nowadays the state of the art in terms of performance~\cite{lecun2015deep}. While the fundamental concepts behind ANNs were introduced in the 1980s, only more recently record-breaking performance has been achieved, thanks to advancements in computational capabilities (especially GPU and HPC) and the advent of the big data era. 

For example, Convolutional Neural Networks (CNNs) achieve unprecedented performance in skin cancer classification~\cite{esteva2017dermatologist}, and architectures based on Recurrent Neural Networks (RNN) are able to decode the electrocorticogram with high accuracy and at natural-speech rates~\cite{makin2020machine}. ANNs are also key components of new-generation autonomous driving systems~\cite{mozaffari2020deep, grigorescu2020survey} and surveillance systems such as Synthetic Aperture Radar (SAR)~\cite{CNN_SAR2016,Martorella2022}. In space-based surveillance, ANNs have the capability of accurately classifying vessels from satellite sensors~\cite{SoldiPartI,SoldiPartII}. In the context of maritime situational awareness and autonomous navigation~\cite{forti_proc_IEEE}, RNNs are able to accurately predict vessel positions several hours ahead~\cite{Capobianco2021}. In video analysis and image understanding, ANNs methods represent the state of the art for object detection~\cite{ObjectDetection_DL_2019} and multi-object tracking~\cite{traktor}. ANNs are also used in Multiple Input Multiple Output (MIMO) communications~\cite{Learning_to_Detect_TSP_2019}, active sensing for communications~\cite{sohrabi2021active}, radar and sonar processing~\cite{DL_Radar_2021,McDonald2021,AES_Magazine2022}.

\begin{figure}
    \centering%
    \includegraphics[trim=5 5 20 20,clip,width=\columnwidth]{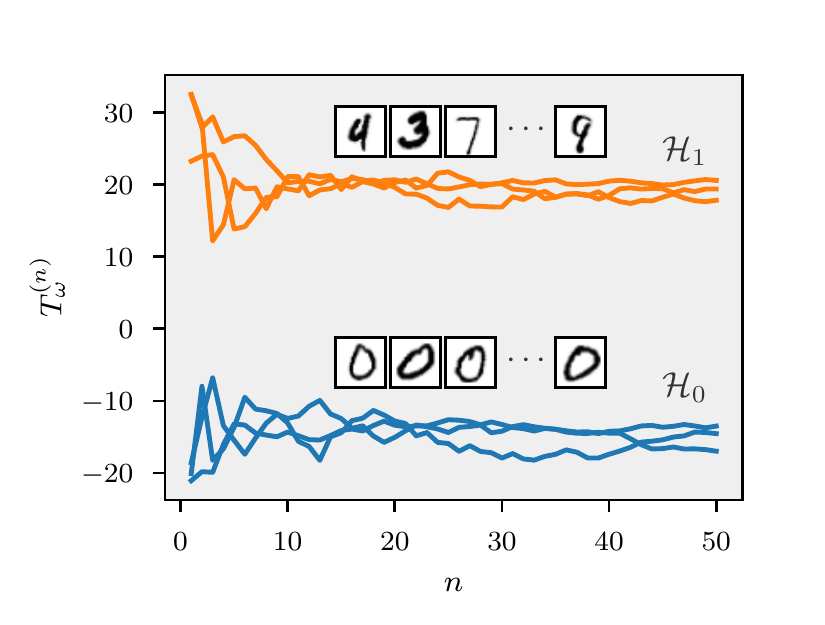}
    \caption{Three different realizations of the decision statistic $T_{\bm{\omega}}^{(n)}$ versus $n$ for each of the two hypotheses ${\cal H}_0$ (digit $0$) and ${\cal H}_1$ (digits $\geq 1$). Each observation $x_i$, $i=1,2,\dots,n$, is an image of a handwritten digit from the MNIST database.}
    \label{fig:d3f_Tw_n}
\end{figure}

Compared to their large success, less is known about the fundamental mathematical properties of ANNs; indeed, ANN methods are usually regarded as ``black boxes''~\cite{alain2017understanding}. Several attempts to fill this gap have been made in recent years. In this paper, we describe two mathematical frameworks to analyse the performance of a generic ML binary classifier in terms of asymptotic error probabilities, and we test them using the popular MNIST (Modified National Institute of Standards and Technology) dataset. The first asymptotic framework is based on the Central Limit Theorem (CLT)~\cite{Lehmann-testing}, and the second is based on the Large Deviations Principle (LDP)~\cite{Dembo-Zeitouni}. They actually both provide approximations of the error probabilities that could be quite accurate even for small values of $n$. More details and some generalizations are available in the extended version of this manuscript~\cite{braca2022statistical}, but here we offer deeper illustration of the results, most particularly involving the MNIST dataset.

\begin{table}
\renewcommand{\arraystretch}{1.3}
\caption{Error probability approximations}
\label{ref:tab}
\centering
\begin{tabular}{r>{\centering}m{0.36\columnwidth}>{\centering\arraybackslash}m{0.36\columnwidth}}
\toprule\toprule
& {\small \textbf{Small deviations}} & {\small \textbf{Large deviations}}\\\midrule
$\alpha_n$ & $ Q\left({\displaystyle\sqrt{n}\ \frac{\gamma - \mu_{0}}{\sigma_{0}}}\right)$                                        & $\zeta_{n,0}(\gamma)\,e^{-n\,I_0(\gamma)}$\\\midrule
$\beta_n$ & $Q\left({\displaystyle \sqrt{n}\ \frac{\mu_{1} - \gamma}{\sigma_{1}}}\right)$    &  $\zeta_{n,1}(\gamma)\,e^{-n\,I_1(\gamma)}$\\        
\bottomrule
\end{tabular}
\end{table}

\section{PROBLEM FORMULATION}
\label{sec:problem}

Consider a family of real-valued decision statistics $T^{(n)}$ that process a sequence of independent and identically distributed (IID) observations ${\cal X}^{(n)} = \left(x_i\right)_{i=1}^{n}$, where $n$ is the number of observed data. The observation $x_i$ can also be an entire image, or another collection of raw data. 
The goal is to decide between two hypotheses
${\cal H}_0$ and ${\cal H}_1$, where the datum $x_i$ is distributed as $f_0(x_i)$ under ${\cal H}_0$, or $f_1(x_i)$ under ${\cal H}_1$, $\forall i=1,\dots,n$.

\begin{figure*}
     \centering
     \begin{subfigure}[b]{0.33\textwidth}
         \centering
         \includegraphics[trim=10 0 20 20,clip,width=\textwidth]{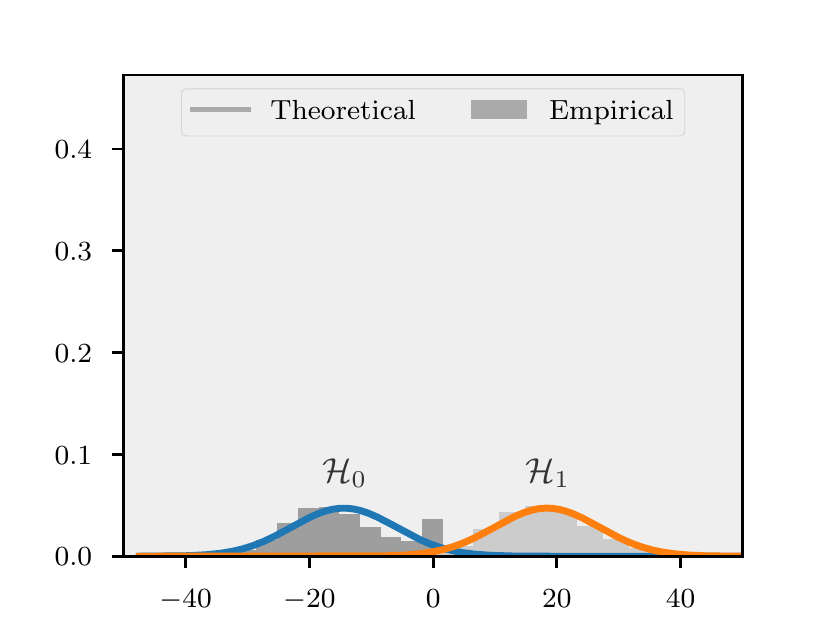}
         \caption{$n=1$}
         \label{fig:d3f_distribution_n_1}
     \end{subfigure}
     \hfill
     \begin{subfigure}[b]{0.33\textwidth}
         \centering
         \includegraphics[trim=10 0 20 20,clip,width=\textwidth]{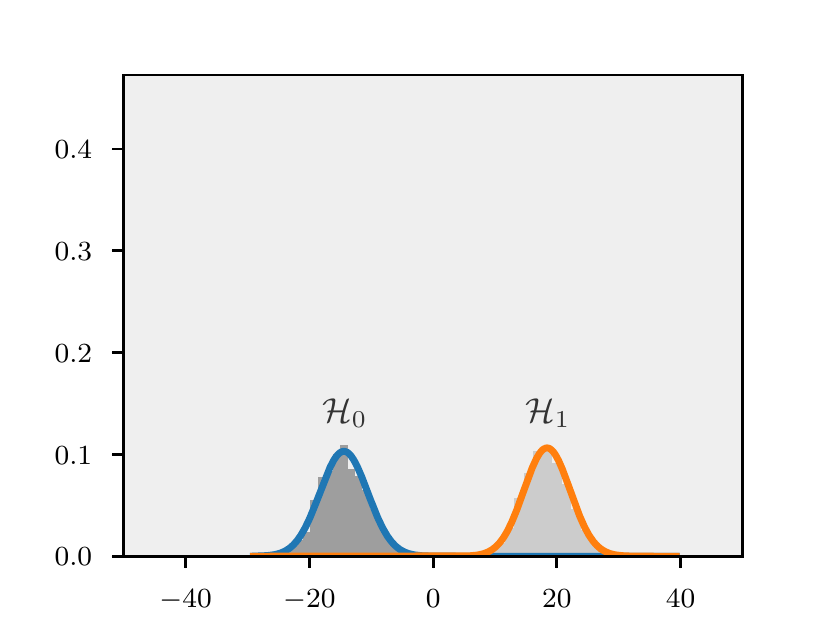}
         \caption{$n=5$}
         \label{fig:d3f_distribution_n_5}
     \end{subfigure}
     \hfill
     \begin{subfigure}[b]{0.33\textwidth}
         \centering
         \includegraphics[trim=10 0 20 20,clip,width=\textwidth]{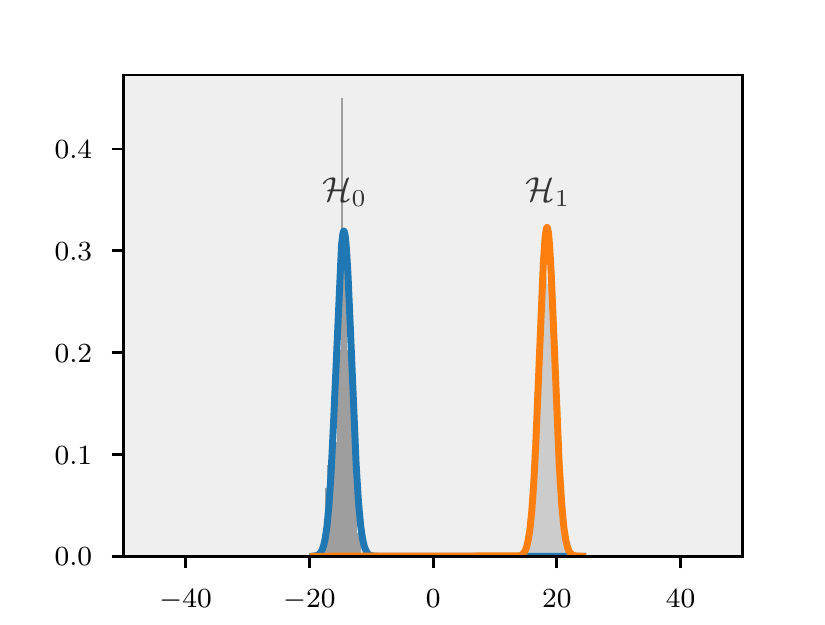}
         \caption{$n=50$}
         \label{fig:d3f_distribution_n_50}
     \end{subfigure}
        \caption{Empirical distributions of the D3F~\eqref{eq:D3F_sum} for different values of $n$. %
        The CLT approximations are Gaussian distributions (solid lines) centered in $\mu_k$ with variance $\sigma_k^2/n$ under ${\cal H}_0$, $k=0,1$.} 
        \label{fig:d3f_distributions}
\end{figure*}

Since the distribution of the observations under ${\cal H}_0$ and ${\cal H}_1$ is often unknown, or too complex to derive, we focus on the case that the decision statistic is provided by a learning mechanism operating on a sufficiently large, finite, labeled, \emph{training set} ${\cal Y}$ available for each hypothesis and independent of ${\cal X}^{(n)}$. The decision statistic $T^{(n)} = T^{(n)}_{\bm{\omega}}$ is referred to as the Data-Driven Decision Function (D3F), characterized by a set $\bm{\omega}$ of parameters that are learned during the training phase. Given the independence of the observations in the sequence ${\cal X}^{(n)}$, it is natural to mimic the structure of the optimal detection statistic, the Log-Likelihood Ratio (LLR), which is the summation of the elementwise LLR $\log\frac{f_1(x_i)}{f_0(x_i)}$ of each $x_i$, $\forall i=1,2,\dots,n$.
As a result, and to pursue a model-based ML strategy, the D3F statistic $T^{(n)}_{\bm{\omega}}$ will be the summation of the elementwise D3F $t_{\bm{\omega}}(x_i)$ in order to  approximate the LLR
\begin{equation}
    T^{(n)}_{\bm{\omega}} = \frac{1}{n} \sum\nolimits_{i=1}^n t_{\bm{\omega}}(x_i).
    \label{eq:D3F_sum}
\end{equation}
Fig.~\ref{fig:d3f_Tw_n} shows three different realizations of $T^{(n)}_{\bm{\omega}}$ as $n$ increases, where $x_i$'s are images from the MNIST database.

We assume that the elementwise D3F $t_{\bm{\omega}}(x_i)$ can be learned from the training set ${\cal Y}$ by standard ML techniques (see details in~\cite{braca2022statistical}). A meaningful choice is to use an ANN trained with the binary cross-entropy loss function and uniform prior (such as in a balanced training set). Then, the elementwise D3F is given by
\begin{equation}
     t_{\bm{\omega}} (x_i) = \log p_{\bm{\omega}}^{({\cal H}_1)}(x_i) - \log p_{\bm{\omega}}^{({\cal H}_0)}(x_i),
\end{equation}
where $p_{\bm{\omega}}^{({\cal H}_k)}$, $k=0,1$, are the outputs of the ANN. As usual, they are interpreted as approximations of the posterior hypothesis probabilities. %
The statistical test is defined as follows:
\begin{equation}
   \left\{ 
   \begin{array}{ll}
     T_{\bm{\omega}}^{(n)} \geq \gamma_n:   & \quad \textnormal{decide ${\cal H}_1$},  \\
     T_{\bm{\omega}}^{(n)} < \gamma_n:  & \quad \textnormal{decide ${\cal H}_0$},
   \end{array} \right.  
   \label{eq:test0}
\end{equation}
where $\gamma_n$ is a decision threshold. The error probabilities of such a test are 
\begin{equation}
    \alpha_n = \mathbb{P} \left[T_{\bm{\omega}}^{(n)} \geq \gamma_n \left| {\cal H}_0 \right. \right], \qquad 
    \beta_n  = \mathbb{P} \left[T_{\bm{\omega}}^{(n)}  < \gamma_n \left| {\cal H}_1 \right. \right].
    \label{eq:error_prob}
\end{equation}

We study the detection performance of $T_{\bm{\omega}}^{(n)}$ when $n$ is large and propose suitable approximations for the finite sample-size regime of $n$. A more general framework, where the input sequence ${\cal X}^{(n)}$ is not necessarily composed by IID observations, is presented in~\cite{braca2022statistical}.

\section{ASYMPTOTIC BEHAVIOUR OF THE D3F}
\label{sec:large_deviations}

For a fixed set of parameters $\bm{\omega}$, such as the network weights in an ANN, the elementwise D3F acts as a transformation of the observed samples, i.e., $\tau_i =  t_{\bm{\omega}}(x_i)$, $\forall i=1,2,\dots,n$, where $\tau_i$ is a real-valued random variable. Consequently, the D3F $T_{\bm{\omega}}^{(n)}$ is the arithmetic mean of $n$ IID real-valued random variables. Given that the parameters $\bm{\omega}$ are fixed, the randomness of $\tau_i$ depends only on $x_i$. 
Exploiting both the CLT~\cite{Lehmann-testing} and the LDP~\cite{Dembo-Zeitouni}, we can establish the two following asymptotic results for the D3F~\eqref{eq:D3F_sum}, see details in~\cite{braca2022statistical}.  

\textbf{Small deviations.} From~\eqref{eq:D3F_sum}, the CLT~\cite{Lehmann-testing} establishes that
the normalized D3F is (asymptotically) distributed as a Gaussian random variable with mean $\mu_k := {\mathbb E}\left[\tau_i  \left| {\cal H}_k \right.\right]$, $k=0,1$, where $ {\mathbb E}\left[ X | {\cal H}_k \right]$ denotes the expected value of $X$ under ${\cal H}_k$. %
Formally, we have the following convergence in distribution
\begin{equation}
    \sqrt{n}(T^{(n)}_{\bm{\omega}}-\mu_k) \stackrel{d}{\longrightarrow} {\cal N} (0,\sigma_k^2),\qquad k=0,1,
    \label{eq:CLT}
\end{equation}
where $\sigma_k$ is the standard deviation of $\tau_i$ (elementwise D3F).

\textbf{Large deviations}. %
The LDP states that the probability of the event that the D3F deviates from the mean
decreases exponentially with $n$. This is the case of the error probabilities~\eqref{eq:error_prob}. Let $\gamma \in (\mu_0,\mu_1)$, the error probabilities of the D3F test~\eqref{eq:test0} with $\gamma_n = \gamma$ satisfies~\cite{braca2022statistical}  
\begin{equation}
\begin{split}
     \lim_{n\rightarrow \infty} \frac{1}{n} \log\alpha_n&= - I_0(\gamma), \\ \lim_{n\rightarrow \infty} \frac{1}{n} \log\beta_n &= - I_1(\gamma),
 \end{split}
     \label{eq:LDP_D3F}
\end{equation}
where $I_k(x)$ is the rate function under ${\cal H}_k$, $k=0,1$, given by the Fenchel-Legendre transform of the Log-Moment Generating Function (LMGF) of the elementwise D3F, $\varphi_{k} (t)$, i.e.,
\begin{equation}
    I_k(x) = \sup_{t \in \mathbb{R}} \left[x\,t - \varphi_k(t) \right],
\end{equation}
with $\varphi_{k} (\eta) = \log {\mathbb E}\left[ {e}^{\eta\,t_{\bm{\omega}}(x_i)} \left| {\cal H}_k\right.\right]$.
The limits in~\eqref{eq:LDP_D3F} can be further refined by computing the most representative sub-exponential terms, based on a refined asymptotic framework referred to as exact asymptotics (see details in~\cite{braca2022statistical}). Then, the error probabilities are approximated as $\zeta_{n,k}(\gamma)\,e^{-n\,I_k(\gamma)}$, with $k=0,1$, where $\zeta_{n,k}$ is given by 
\begin{equation}
    \zeta_{n,k} =\left( t_{\gamma,k}\sqrt{2\pi n \varphi_k^{\prime\prime}\left( (-1)^k\,t_{\gamma,k} \right)}\right)^{-1},
\end{equation}
and $t_{\gamma,k}:\varphi_k^\prime\left( (-1)^k\, t_{\gamma,k}\right)  = \gamma.$ 

The CLT and the LDP offer two different approximations for the error probabilities~\eqref{eq:error_prob}, reported in Table~\ref{ref:tab}. The CLT approximation stems from~\eqref{eq:CLT}, where %
$T^{(n)}_{\bm{\omega}}$ is approximated as a Gaussian distribution centered in $\mu_k$ with variance $\sigma_k^2 / n$, while the LDP one follows from the exact asymptotics.

Generally, not all the parameters necessary for the small and large deviations approximations
are available, even when the distributions of data are known in advance. In our context, such parameters must be estimated from the available data, referred to as the characterization  set, which can include the training set~\cite{braca2022statistical}. The basic idea is to replace the expectations to compute the moments and the LMGF of the elementwise D3F with their related sample means. As elaborated in~\cite{braca2022statistical}, when the number of samples in the characterization set is large enough, the estimated parameters will be accurate enough to compute the approximate error probability curves. %

\section{COMPUTER EXPERIMENTS AND RESULTS}
\label{sec:experiments}
In this section, the theory summarized above and detailed in~\cite{braca2022statistical} is employed to analyze the performance of a D3F test using the popular MNIST dataset, a database of handwritten digits containing \num{60000} training and \num{10000} testing grayscale images~\cite{mnist}. All the images are normalized and centered in a fixed-size image with $28\times 28$ pixels; the dimensionality of each image is thus $28^2=784$. 

The statistical testing problem is to decide if a sequence of $n$ observations all come from ${\cal H}_0$ (digit $0$) or ${\cal H}_1$  (all digits $\geq 1$). Contextualized to the MNIST case, the single observation $x_i$ is an image realization of a digit $\theta_i$ and $n$ represents the number of images in the sequence. Under ${\cal H}_0$, $\theta_i \in \left\{0\right\}$, while under ${\cal H}_1$, $\theta_i \in \left\{1,2,\dots,9\right\}$, where $\theta_i$ is randomly extracted for each $i=1,2,\dots,n$ (see a pictorial representation of such sequences in Fig.~\ref{fig:d3f_Tw_n}). Each observation $x_i$ is processed according to the D3F~\eqref{eq:D3F_sum}, and the final decision is taken according to~\eqref{eq:test0}. The elementwise D3F is a fully-connected neural network with a single hidden layer of $8$ units and ReLU activation function, trained with Adam~\cite{kingma2014adam} to minimize the binary cross entropy loss for a fixed number of epochs.

\begin{figure}
    \centering%
    \vspace{-2em}%
    \begin{subfigure}[t]{0.48\textwidth}
        \centering
        \includegraphics[width=\textwidth,trim=5 5 15 0,clip]{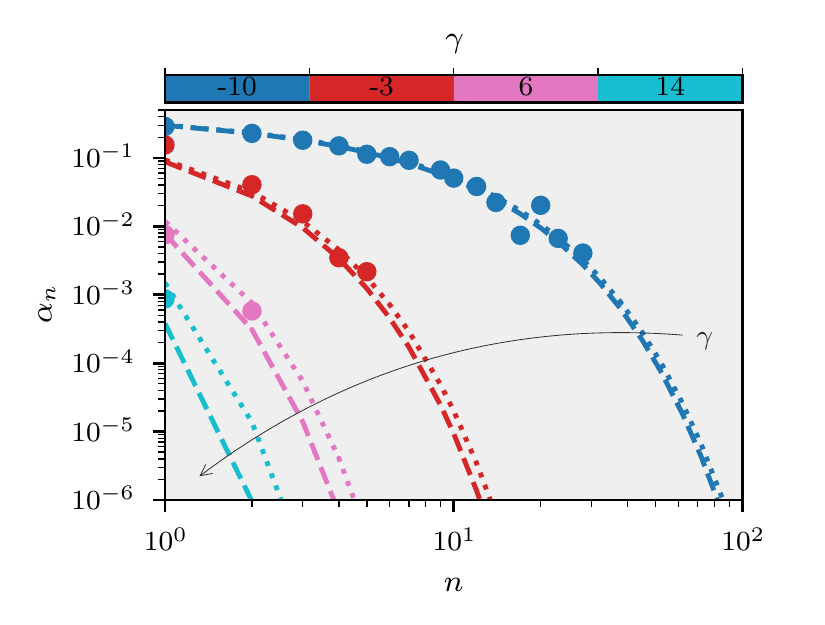}
    \end{subfigure}
    \\
    \begin{subfigure}[t]{0.48\textwidth}
        \centering
        \includegraphics[width=\textwidth,trim=5 5 15 31,clip]{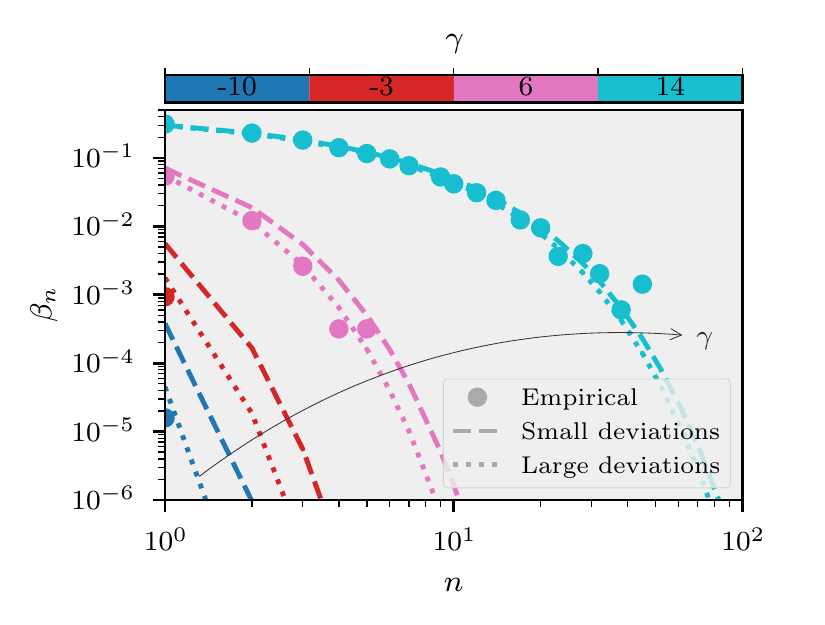}
    \end{subfigure}%
    \\[-0.5em]%
    \begin{subfigure}[t]{0.48\textwidth}
        \centering
        \includegraphics[width=\columnwidth,trim=5 5 15 15,clip]{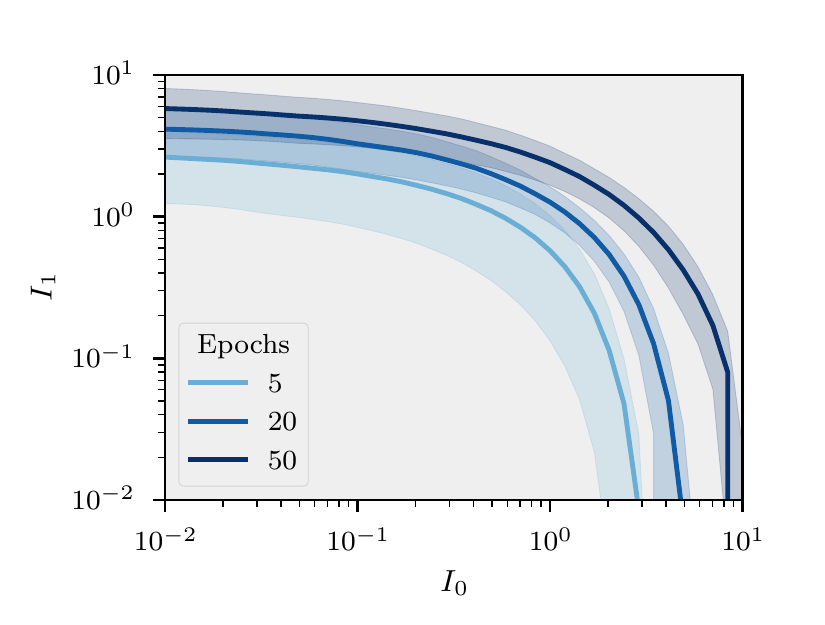}
    \end{subfigure}
    \vspace{-2em}%
    \caption{Top and middle panel: error probabilities $\alpha_n$ (top) and $\beta_n$ (middle) versus $n$, the number of observations (images of the MNIST database). Each curve refers to a different value of $\gamma$, identified by different colors. For a given threshold we compare the empirical error probabilities (circles) with their small deviation (dashed curves) and large deviations (dotted curves) approximations. Bottom panel: probability error rates $I_0(\gamma)$ vs $I_1(\gamma)$, with variable $\gamma \in [\mu_0, \mu_1]$. Different curves are related to different realizations of the ANN parameters $\bm{\omega}$ increasing the training epochs. The shaded bands represent the variability ($\pm 1$ standard deviation) of the rates induced by $\bm{\omega}$.}
    \label{fig:d3f_aio_fig}
\end{figure}

In Fig.~\ref{fig:d3f_Tw_n} we show %
three different realizations of the decision statistic, under ${\cal H}_0$ and ${\cal H}_1$, as $n$ increases. The decision statistics converge to their expected values, i.e., $\mu_0$ ($\approx -15$) under ${\cal H}_0$ and $\mu_1$ ($\approx 20$) under ${\cal H}_1$; note that the expected values $\mu_0$ and $\mu_1$ can vary significantly not only depending on the training strategy and its parameters, but also from  %
run to run because of the randomness of the stochastic gradient descent. %
Such randomness is captured by $\bm{\omega}$, and all the convergences that we study are conditioned to a realization of $\bm{\omega}$.   

The convergence behaviour is aligned with the predictions of the CLT~\eqref{eq:CLT}, and indeed the empirical histograms of the decision statistics in Fig.~\ref{fig:d3f_distributions} are very close to the theoretical Gaussian distributions (solid curves) centered in $\mu_k$ with variance $\sigma_k^2/n$ that decreases linearly with the number of observations. It is evident a very good agreement between the empirical histograms and the theoretical distributions even for small values of $n$. From the behaviour of the decision statistic under the two hypotheses, it follows that we can set the threshold between $\mu_0$ and $\mu_1$ to have both the error probabilities vanishing with $n$ as expected from the large deviations result~\eqref{eq:LDP_D3F}.
The top and middle panel of Fig.~\ref{fig:d3f_aio_fig} illustrate the empirical error probabilities and their approximations (summarized in Tab.~\ref{ref:tab}) stemming from the CLT and the LDP for several values of the test threshold $\gamma_n=\gamma$. It can be observed that both the approximations are quite close to the empirical error curves. However--as expected--the best agreement is given by the LDP approximation (see also the discussion in~\cite{braca2022statistical}). 

Given that the MNIST dataset has a fixed number of samples %
for each digit, the empirical curves are computed with less and less realizations as $n$ increases. For this reason, we are able to compute reliable empirical error probabilities lower than (or equal to) $10^{-4}$ only for $n=1$ under ${\cal H}_1$, where we have in total \num{63097} samples (\num{54077} from the training set and  \num{9020} from the testing set), and we are able to compute an empirical error probability around $10^{-3}$ only for $n=1$ under ${\cal H}_0$ where we have in total \num{6903} samples (\num{5923} from the training set and  \num{980} from the testing set). Note that it is possible to use the training set to estimate the relevant LDP and CLT parameters (see~\cite{braca2022statistical}).

In Fig.~\ref{fig:d3f_aio_fig}, top and middle panels, it is possible to observe that the higher (lower) $\gamma$, the faster $\alpha_n$ ($\beta_n$) vanishes. This behavioural trade-off is intuitive and present in all the detection problems. Indeed, the classic Receiver Operating Characteristic (true positive rate $1-\beta_n$ against the false positive rate $\alpha_n$) is a concave curve, obtained by varying the threshold. By increasing the threshold the false positive rate decreases, as well as does the true positive rate~\cite{Lehmann-testing}. This behavioural trade-off is also present in the rate functions $I_0(\gamma)$ and $I_1(\gamma)$, reported in the bottom panel of Fig.~\ref{fig:d3f_aio_fig}, which rule the rate of convergence to zero of the error probabilities. Specifically, we have $I_k(\gamma) = 0$ when $\gamma = \mu_k$. Assuming $\mu_0 < \mu_1$, $I_0(\gamma)$ (or $I_1(\gamma)$) increases (or decreases) with $\gamma$ reaching its maximum at $\gamma = \mu_1$ ($\gamma = \mu_0$). In the bottom panel of Fig.~\ref{fig:d3f_aio_fig} we report different curves averaged over different realizations of the network parameters $\bm{\omega}$ increasing the training epochs, all the other parameters being equal. Clearly, it is expected that using more training epochs allow the gradient descent to converge to a better solution, thus leading to better performance (at least until a saturation point is reached). This is indeed reflected in Fig.~\ref{fig:d3f_aio_fig}, bottom panel, where models trained for more epochs achieve on average higher rates $I_1$ at the same rate $I_0$.

\section{Conclusion}
\label{sec:conclusion}
In this paper, we have described a novel method to analyze the performance of a generic machine learning binary classifier based on the large deviations theory. The classifier relies on a suitable decision statistic, referred to as the data-driven decision function, which is learned from training data, and its performance is defined in terms of error probabilities and their convergence rates.  
We have described the conditions that lead the data-driven decision function to exhibit error probabilities vanishing exponentially with the number of observations $n$. Two approximations for the error probabilities are proposed based on the central limit theorem and the large deviations principle. The validity of the proposed approach has been validated using a fully connected neural network-based classifier and the MNIST database.

\bibliographystyle{IEEEtran}
\bibliography{IEEEabrv,refs}

\end{document}